\def\year{2021}\relax
\documentclass[letterpaper]{article} 
\usepackage{aaai21}  
\usepackage{times}  
\usepackage{helvet} 
\usepackage{courier}  
\usepackage[hyphens]{url}  
\usepackage{graphicx} 
\usepackage{graphicx}  
\usepackage{natbib}  
\usepackage{caption} 
\usepackage{multirow}
\newcommand{\tabincell}[2]{\begin{tabular}{@{}#1@{}}#2\end{tabular}}
\usepackage{amsmath,bm}
\usepackage{dsfont}
\usepackage{color} 
\usepackage[switch]{lineno}  %

\title{Voxel R-CNN: Towards High Performance Voxel-based 3D Object Detection}
\author{
	Jiajun Deng\textsuperscript{\rm 1}, 
	Shaoshuai Shi\textsuperscript{\rm 2}, 
	Peiwei Li\textsuperscript{\rm 1}, 
	Wengang Zhou\textsuperscript{\rm 1,3}, 
	Yanyong Zhang\textsuperscript{\rm 4}, 
	Houqiang Li\textsuperscript{\rm 1,3} \\
}

\affiliations{
	\textsuperscript{\rm 1} CAS Key Laboratory of GIPAS, EEIS Department, University of Science and Technology of China\\
	\textsuperscript{\rm 2} Multimedia Laboratory, The Chinese University of Hong Kong \\
	\textsuperscript{\rm 3} Institute of Artificial Intelligence, Hefei Comprehensive National Science Center \\
	\textsuperscript{\rm 4} Department of Computer Science, University of Science and Technology of China\\
	 \{dengjj, lpw0622\}@mail.ustc.edu.cn, \{ssshi\}@ee.cuhk.edu.hk \{zhwg, yanyongz, lihq\}@ustc.edu.cn
}

\begin{document}
	\maketitle
	
	\begin{abstract}

		Recent advances on 3D object detection heavily rely on how the 3D data are represented, \emph{i.e.}, voxel-based or point-based representation. Many existing
		high performance 3D detectors are point-based because this structure can better retain precise point positions. Nevertheless, point-level features lead to high computation overheads due to unordered storage. In contrast, the voxel-based structure is better suited for feature extraction but often yields lower accuracy because the input data are divided into grids. In this paper, we take a slightly different viewpoint --- we find that precise positioning of raw points is not essential for high performance 3D object detection and that the coarse voxel granularity can also offer sufficient detection accuracy. Bearing this view in mind, we devise a simple but effective voxel-based framework, named Voxel R-CNN. By taking full advantage of voxel features in a two stage approach, our method achieves comparable detection accuracy with state-of-the-art point-based models, but at a fraction of the computation cost.  Voxel R-CNN consists of a 3D backbone network, a 2D bird-eye-view (BEV) Region Proposal Network and a detect head. A voxel RoI pooling is devised to extract RoI features directly from voxel features for further refinement. Extensive experiments are conducted on the widely used KITTI Dataset and the more recent Waymo Open Dataset.  Our results show that compared to existing voxel-based methods, Voxel R-CNN delivers a higher detection accuracy while maintaining a real-time frame processing rate,  \emph{i.e}., at a speed of 25 FPS on an NVIDIA RTX 2080 Ti GPU.  The code is available at \url{https://github.com/djiajunustc/Voxel-R-CNN}.

	\end{abstract}

\section{Introduction}
3D object detection using point clouds has received substantial attention in autonomous vehicles, robotics and augmented/virtual reality. Although the recent development of deep learning has led to the surge of object detection with 2D images \shortcite{ren2015faster,liu2016ssd,redmon2017yolo9000,szegedy2017inception}, it is still non-trivial to apply these 2D methods to 3D point clouds, especially when dealing with the sparsity and unstructured property of point clouds. Besides, the applications usually demand high efficiency from detection systems, making it even harder to design a 3D detector due to the much larger 3D space.

\begin{figure}[!tb]
	\centering {\includegraphics[width=0.47\textwidth]{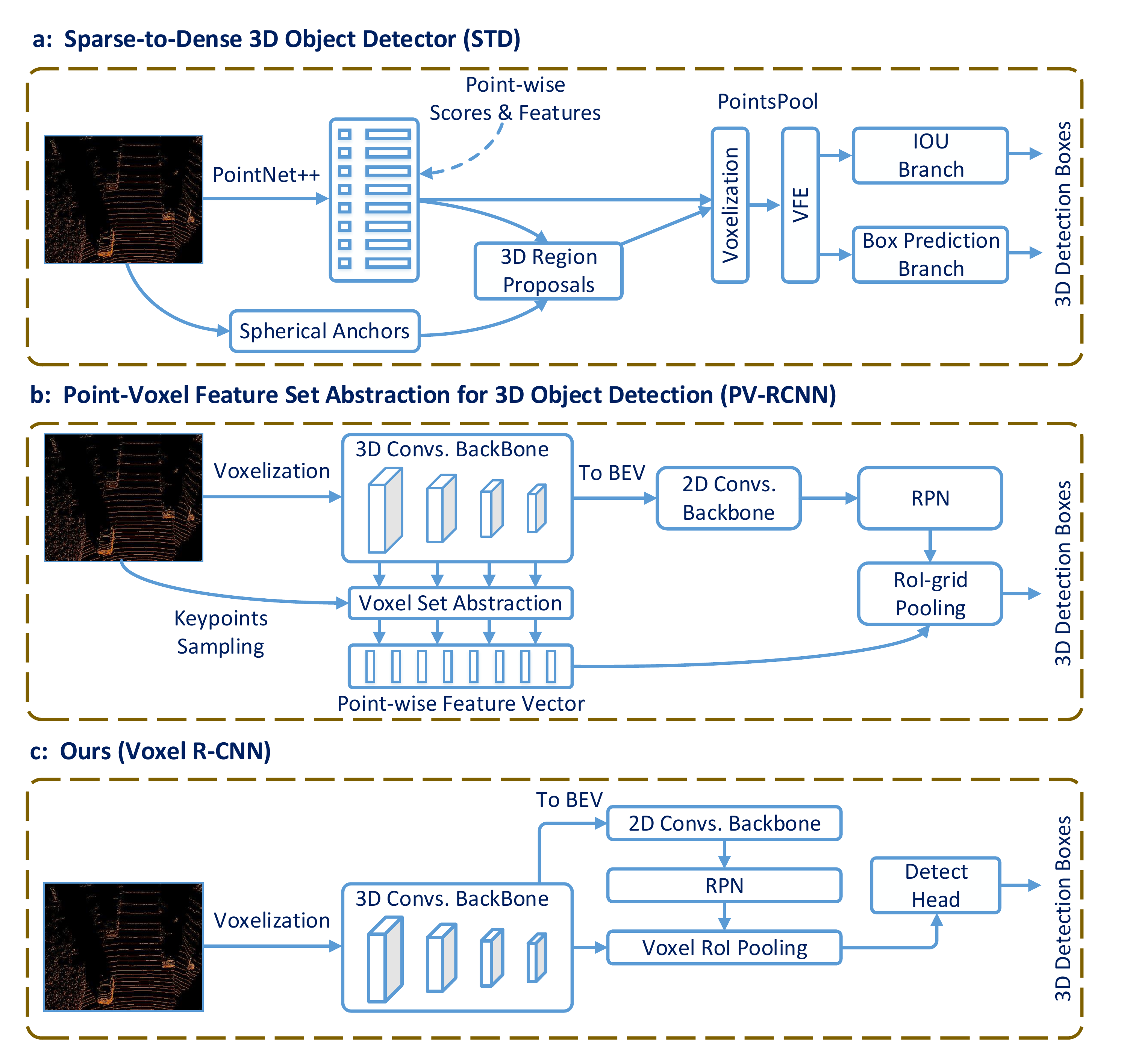}}
	\caption{\small Comparison with state-of-the-art 3D object detection methods in two stage paradigm. Instead of aggregating RoI features from a set of point representations, our method directly extract RoI features from 3D voxel feature volumes.
	}
	\label{fig:intro}
\end{figure}

Existing 3D detection methods can be broadly grouped into two categories, \emph{i.e.}, voxel-based and point-based. The \textit{voxel-based} methods \shortcite{zhou2018voxelnet,yan2018second,lang2019pointpillars} divide point clouds into regular grids, which are more applicable for convolutional neural networks (CNNs) and more efficient for feature extraction due to superior memory locality. Nevertheless, the downside is that voxelization  often causes loss of precise position information. 
Current state-of-the-art 3D detectors are mainly \textit{point-based}, which take raw point clouds as input, and abstract a set of point representations with iterative sampling and grouping \shortcite{qi2017pointnet,qi2017pointnet++}.  The advanced point-based methods \shortcite{Yang_2019_ICCV,shi2019pointrcnn,Yang_2020_CVPR,Shi_2020_CVPR} are top ranked on various benchmarks \shortcite{geigerwe,caesar2020nuscenes,sun2020scalability}. This has thus led to the popular viewpoint that the precise position information in raw point clouds is crucial for accurate object localization.	Despite the superior detection accuracy, point-based methods are in general less efficient because it is more costly to search nearest neighbor with the point representation for point set abstraction. 

As the detection algorithms become mature, we are ready to deploy these algorithms on realistic systems. Here, a new challenge arises: can we devise a method that is as accurate as advanced point-based methods and as fast as voxel-based methods?
In this work, towards this objective, we take the voxel-based framework and try to boost its accuracy. We first argue that precise positioning of  raw point clouds is nice but unnecessary. We observe that voxel-based methods generally perform object detection on the bird-eye-view (BEV) representation, even if the input data are in 3D voxels \shortcite{yan2018second}. 
In contrast, point-based methods commonly rely on abstracted point representations to restore 3D structure context and make further refinement based on the point-wise features, as in Figure \ref{fig:intro} (a) (b).
By taking a close look at the underlying mechanisms, we find that the key disadvantage of existing voxel-based methods stems from the fact that they convert 3D feature volumes into BEV representations without ever restoring the 3D structure context.

Bearing this in mind, we propose to aggregate 3D structure context from 3D feature volumes. Specifically, we introduce a novel voxel-based detector, \emph{i.e.}, Voxel R-CNN, to take full advantage of voxel features in a two stage pipeline (see Figure~\ref{fig:intro} (c)). Voxel R-CNN consists of three core modules: (1) a 3D backbone network, (2) a 2D backbone network followed by the Region Proposal Network (RPN), and (3) a detect head with a new voxel RoI pooling operation. The 3D backbone network gradually abstracts voxels into 3D feature volumes. Dense region proposals are generated by the 2D backbone and the RPN. Then, the RoI features are directly extracted from 3D feature volumes with voxel RoI pooling. In designing voxel RoI pooling, we take the neighbor-aware property (which facilitates better memory locality) to extract neighboring voxel features and devise a local feature aggregation module for further acceleration. Finally, the 3D RoI features are taken for further box refinement.

The main contribution of this work stems from the design of Voxel R-CNN, which strikes a careful balance between accuracy and efficiency. The encouraging experiment results of Voxel R-CNN also confirm our viewpoint: the precise positioning of raw points is not essential for high performance 3D object detection and coarser voxel granularity can also offer sufficient spatial context cues for this task. Please note that our Voxel R-CNN framework serves as a simple but effective baseline facilitating further investigation and downstream tasks.

\section{Reflection on 3D Object Detection}

In this section, we first revisit two representative baseline methods, \emph{i.e.}, SECOND \shortcite{yan2018second} and PV-RCNN \shortcite{Shi_2020_CVPR},  and then investigate the key factors of developing a high performance 3D object detector. 

\subsection{Revisiting}

\subsubsection{SECOND.} \hskip -5pt  SECOND \shortcite{yan2018second} is a voxel-based one stage object detector. 
It feeds the voxelized data to a 3D backbone network for feature extraction. The 3D feature volumes are then converted to BEV representations. Finally, a 2D backbone followed by a Region Proposal Network (RPN) is applied to perform detection. 

\subsubsection{PV-RCNN.} \hskip -5pt PV-RCNN \shortcite{Shi_2020_CVPR} extends SECOND by adding a keypoints branch to preserve 3D structural information.
Voxel Set Abstraction (VSA) is introduced to integrate multiscale 3D voxels features into keypoints. The features of each 3D region proposals are further extracted from the keypoints through RoI-grid pooling for box refinement.

\begin{table}[t]
	\small 
	\begin{center}
		\scalebox{0.99}{
			\setlength\tabcolsep{6pt}
			\begin{tabular}{l|ccc}
				\hline 
				\multirow{2}{*}{\tabincell{c}{Methods}} & 
				\multicolumn{3}{c}{$\text{AP}_\text{3D}$ (\%)} \\

				& Easy & Mod.  & Hard\\
				\hline
				PV-RCNN~\shortcite{Shi_2020_CVPR}& 89.35 & 83.69 & 78.70  \\
				\hline
				SECOND~\shortcite{yan2018second} & 88.61 & 78.62 & 77.22 \\ 
				SECOND +BEV detect head& 89.50 & 79.26 & 78.45\\ 
				\hline
			\end{tabular}
		}
	\end{center}
	\caption{Performance comparison for adding BEV detect head on the top of SECOND. These results are evaluated on the KITTI \textit{val} set with average precision calculated by 11 recall positions for car class.}
	\label{tab:second_with_bev_head}
\end{table}

\begin{table}[t]
	\centering 
	\footnotesize
	\setlength\tabcolsep{3pt}
	\begin{tabular}{|c|c|c|c|c|}
		\hline
		Components & 3D backbone & 2D backbone & VSA & Detect head \\
		\hline
		Avg. time (ms)& 21.1 &  9.4 & 49.4 & 19.1 \\
		\hline
	\end{tabular}
	\caption{Running time comparison for each component in PV-RCNN. This result is calculated by the average over the 3,769 samples in the KITTI \textit{val} set.}
	\label{tab:pvrcnn_timetable}
\end{table}

\subsection{Analysis} 
There exists a large gap between SECOND and PV-RCNN in terms of the detection performance (\emph{i.e.}, accuracy and efficiency). These two methods differ in the following ways. Firstly, SECOND is a one stage method while PV-RCNN takes detect head for box refinement. Secondly,  keypoints in PV-RCNN preserve the 3D structure information, while SECOND performs detection directly on BEV representations. To verify the influence of box refinement and 3D structure information on detection performance, we add a detect head on the top of the 2D backbone network in SECOND. Since the BEV boxes are not aligned with the axis, we exploit Rotated RoI Align for RoI feature extraction. 

\begin{figure*}[t]
	\centering {\includegraphics[width=0.9\textwidth]{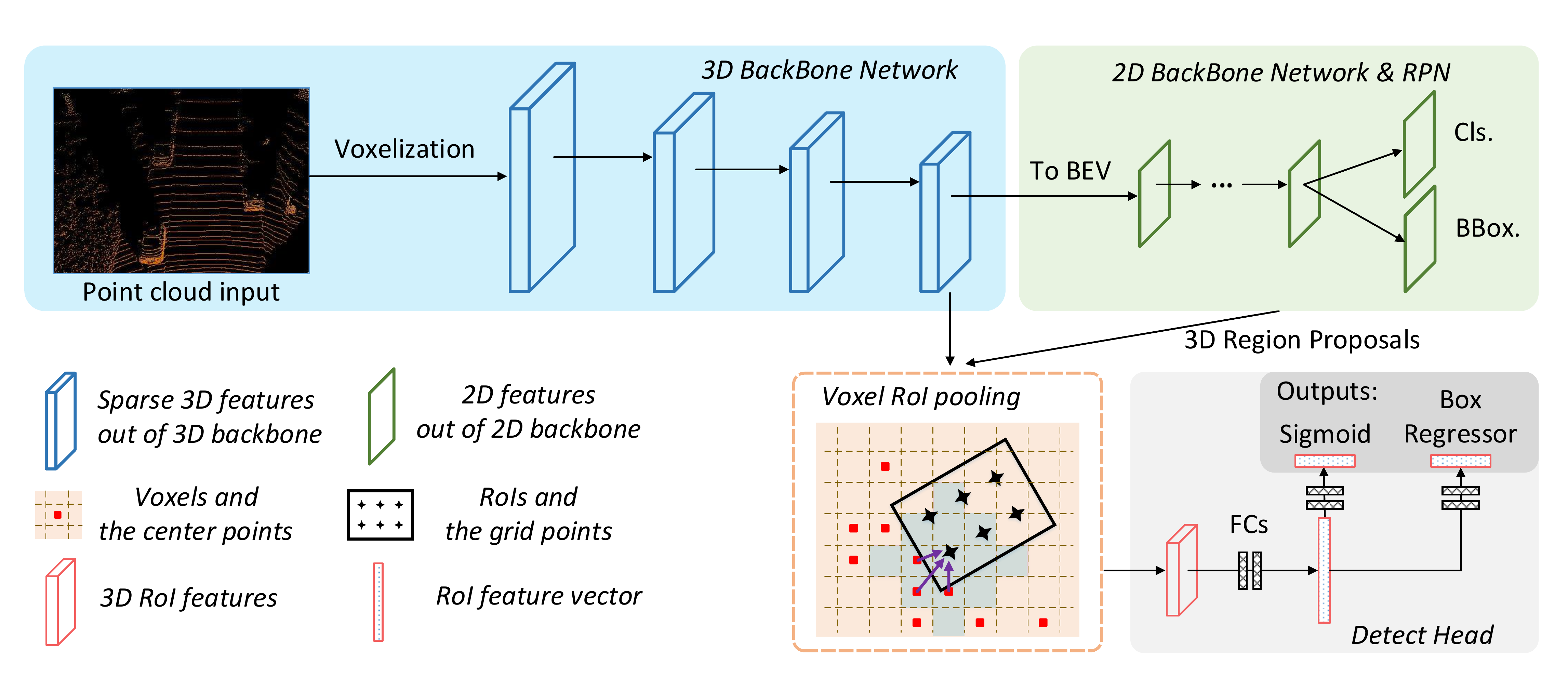}}
	\caption{An overview of Voxel R-CNN for 3D object detection. The point clouds are first divided into regular voxels and fed into the 3D backbone network for feature extraction. Then, the 3D feture volumes are converted into BEV representation, on which we apply the 2D backbone and RPN for region proposal generation. Subsequently, voxel RoI pooling directly extracts RoI features from the 3D feature volumes. Finally the RoI features are exploited in the detect head for further box refinement.
	}
	\label{fig:framework}
\end{figure*}

As illustrated in Table \ref{tab:second_with_bev_head}, directly adding a BEV detect head on top of BEV features leads to 0.6\% AP improvement  for KITTI car moderate data, but has still trailed the accuracy of PV-RCNN thus far. This verifies the effectiveness of box refinement, and also demonstrates that \textit{the capacity of BEV representation is rather limited}. Typically, PV-RCNN integrates voxel features into sampled keypoints with Voxel Set Abstraction. The keypoints works as an intermediate feature representation to effectively preserve 3D structure information. However, as illustrated in Table \ref{tab:pvrcnn_timetable}, \textit{ the point-voxel interaction takes almost half of the overall running time}, which makes PV-RCNN much slower than SECOND.

\subsubsection{Summary.} \hskip -5pt In summary, by analyzing the limitations of bird-eye-view (BEV) feature representations in SECOND and the computation cost of each component  in PV-RCNN, we observe the following: (a) the \textbf{\textit{3D structure is of significant importance}} for 3D object detectors, since the BEV representation alone is insufficient to precisely predict bounding boxes in a 3D space; and (b) \textbf{\textit{the point-voxel feature interaction is time-consuming}} and affects the detector's efficiency. 
These observations motivate us to directly leverage the 3D voxel tensors and develop a voxel-only 3D object detector.

	\section{Voxel R-CNN Design}
	
	In this section, we present the design of Voxel R-CNN, a voxel-based two stage framework for 3D object detection. As shown in Figure \ref{fig:framework}, Voxel R-CNN includes: (a) a 3D backbone network, (b) a 2D backbone network followed by the Region Proposal Network (RPN), and (c) a voxel RoI pooling and a detect subnet for box refinement.  In Voxel R-CNN, we first divide the raw point cloud into regular voxels and utilize the 3D backbone network for feature extraction. We then convert the sparse 3D voxels into BEV representations, on which we apply the 2D backbone network and the RPN to generate 3D region proposals. Subsequently, we use Voxel RoI pooling to extract RoI features, which are fed into the detect subnet for box refinement. Below we discuss these modules in detail. Since our innovation mainly lies in voxel RoI pooling, we discuss it first. 
	
		\begin{figure}[t]
		\centering {\includegraphics[width=0.45\textwidth]{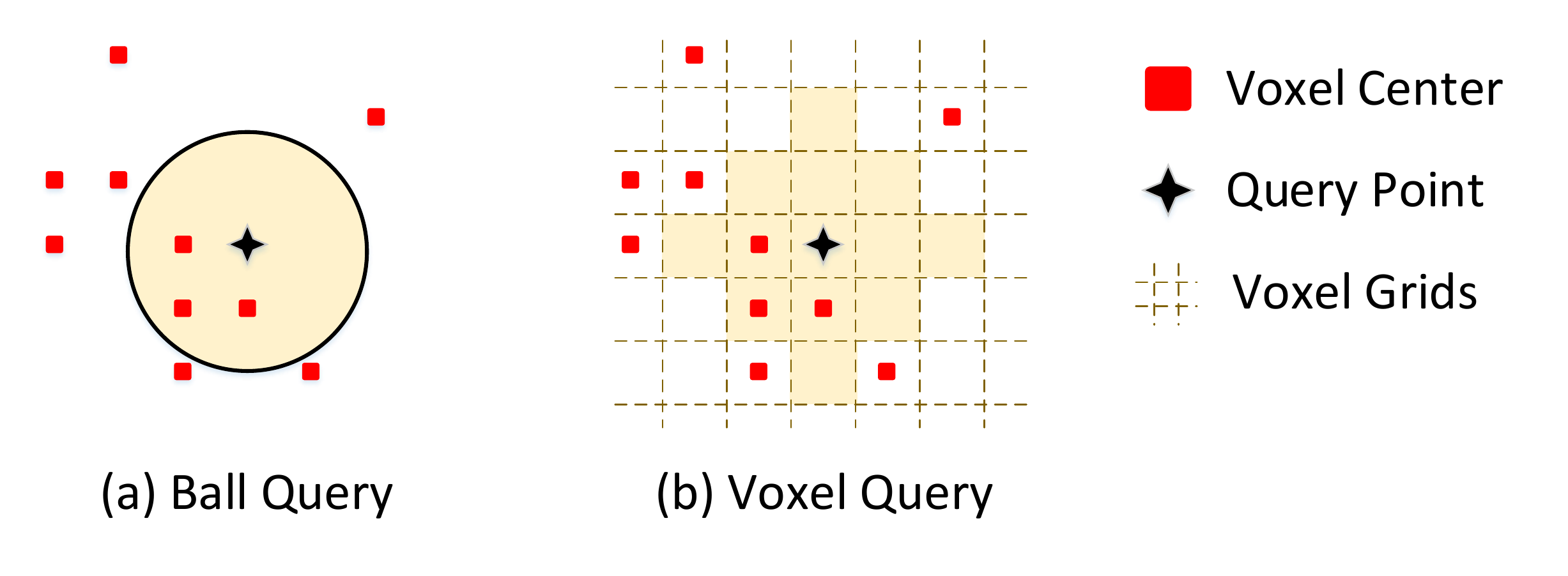}}
		\caption{ Illustration of ball query and our voxel query (performed in 3D space but shown in 2D space here)
		}
		\label{fig:voxel_query}
	\end{figure}

	\subsection{Voxel RoI pooling} 
	To directly aggregate spatial context from 3D voxel feature volumes, we propose voxel RoI pooling.
	
	\subsubsection{Voxel Volumes as Points.} \hskip -5pt  We represent the sparse 3D volumes as a set of non-empty voxel center points $\{\bm{v}_i=(x_i,y_i,z_i)\}_{i=1}^N$  and their corresponding feature vectors $\{\bm{\phi}_i\}_{i=1}^N$. Specifically, the 3D coordinates of voxel centers are calculated with the indices, voxel sizes and the point cloud boundaries.

	\subsubsection{Voxel Query.} \hskip -5pt We propose a new operation, named voxel query, to find neighbor voxels from the 3D feature volumes. 
	Compared to the unordered point clouds, the voxels are regularly arranged in the quantified space, lending itself to easy neighbor access. For example, the 26-neighbor voxels of a query voxel can be easily computed by adding a triplet of offsets $(\Delta_i, \Delta_j, \Delta_k), \Delta_i,\Delta_j,\Delta_k\in\{-1, 0, 1\}$ on the voxel indices $(i, j, k)$. By taking advantage of this property, we devise voxel query to efficiently group voxels. The voxel query is illustrated in Figure \ref{fig:voxel_query}. The query point is first  quantified into a voxel, and then the neighbor voxels can be efficiently obtained by indices translation. 
	We exploit Manhattan distance in voxel query and sample up to $K$ voxels within a distance threshold. Specifically, the Manhattan distance $D(\alpha,\beta)$ between the voxel $\alpha=(i_\alpha,j_\alpha,k_\alpha)$ and $\beta=(i_\beta,j_\beta,k_\beta)$ is computed as:
	\begin{equation}\label{equ:distance}
		D_m(\alpha,\beta)=|i_{\alpha} - i_{\beta}| + |j_{\alpha} - j_{\beta}| + |k_{\alpha} - k_{\beta}|.
	\end{equation}
	
	Suppose there are $N$ non-empty voxels in the 3D feature volumes and we utilize ball query to find neighboring voxels to a given query point, the time complexity is $O(N)$. Nevertheless, the time complexity of conducting voxel query is only $O(K)$, where K is number of neighbors. The neighbor-aware property makes it more efficient to group neighbor voxel features with our voxel query than to group neighbor point features with ball query \shortcite{qi2017pointnet++}.
	
	\subsubsection{Voxel RoI Pooling Layer.} \hskip -5pt 
	We design the voxel RoI pooling layer as follows. It starts by dividing a region proposal into $G\times G\times G$ regular sub-voxels. The center point is taken as the grid point of the corresponding sub-voxel. Since 3D feature volumes are extremely sparse (non-empty voxels account for $<3\%$ spaces), we cannot directly utilize max pooling over features of each sub-voxel as in \shortcite{girshick2015fast}. Instead, we integrate features from neighboring voxels into the grid points for feature extraction. Specifically, given a grid point $\bm{g}_i$, we first exploit voxel query to group a set of neighboring voxels $\Gamma_i=\{\bm{v}_i^1, \bm{v}_i^2,\cdots,\bm{v}_i^K\}$. Then, we aggregate the neighboring voxel features with a PointNet module \shortcite{qi2017pointnet} as:
	\begin{equation}\label{equ:mlpmaxpool}
		\bm{\eta}_i=\max \limits_{k=1,2,\cdots,K}\{\Psi([\bm{v}_i^k-\bm{g}_i;\bm{\phi}_i^k])\},
	\end{equation}
	where $\bm{v}_i - \bm{g}_i$ represents the relative coordinates, $\bm{\phi}_i^k$ is the voxel feature of $\bm{v}_i^k$, and $\Psi(\cdot)$ indicates an MLP. The max pooling operation $\max(\cdot)$ is performed along the channels to obtain the aggregated feature vector $\bm{\eta}_i$. Particularly, we exploit Voxel RoI pooling to extract voxel features from the 3D feature volumes out of the last two stages in the 3D backbone network. And for each stage, two Manhattan distance thresholds are set to group voxels with multiple scales. Then, we concatenate the aggregated features pooled from different stages and scales to obtain the RoI features.
		
		\begin{figure}[t]
		\centering {\includegraphics[width=0.48\textwidth]{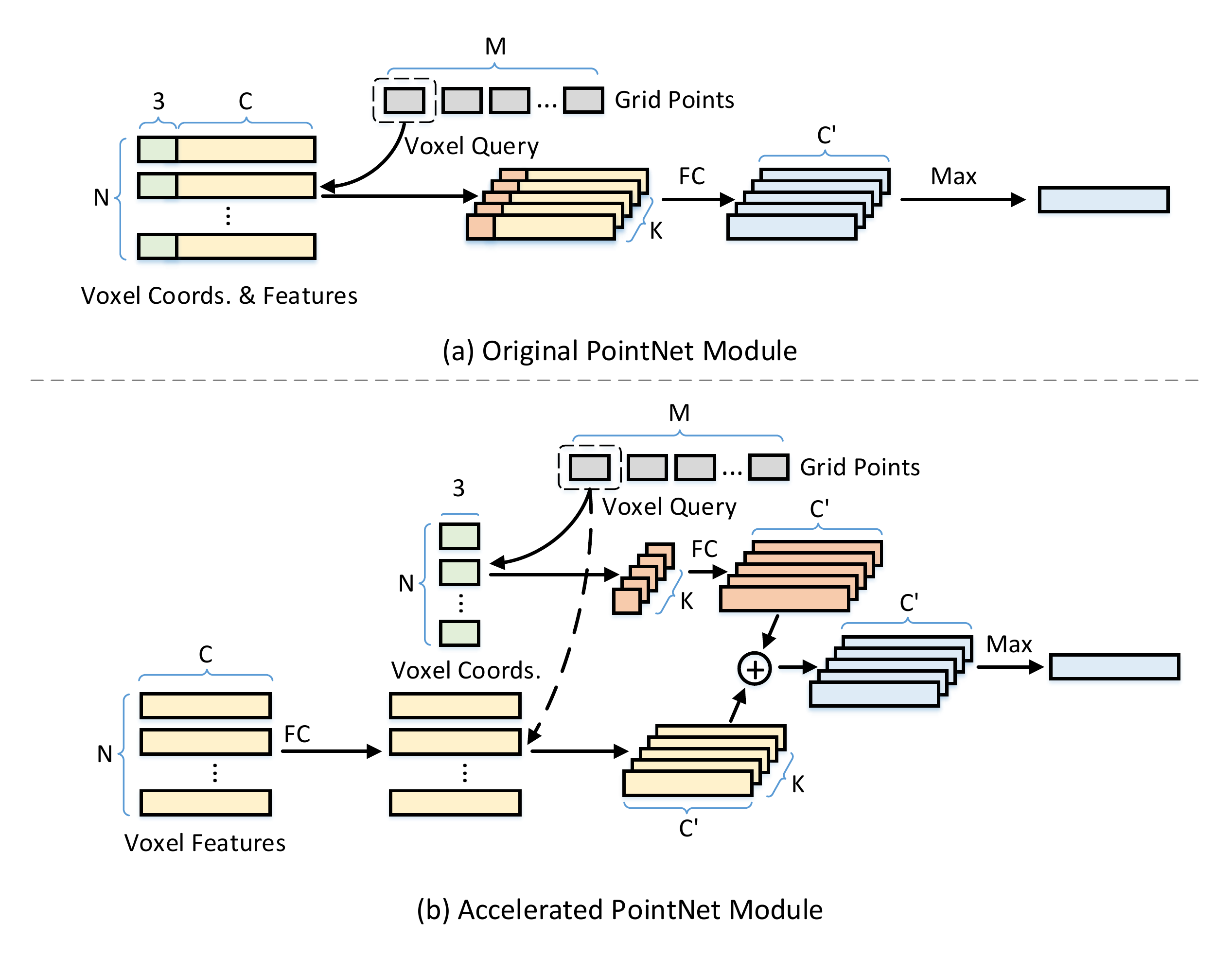}}
		\caption{Illustration of different schemes to aggregate voxel features: (a) original PointNet module; (b) accelerated PointNet module. We only pick one grid point as an example, and the same operations are applied on all the other grid points.}
		\label{fig:mlp}
	\end{figure}

	\subsubsection{Accelerated Local Aggregation.} \hskip -5pt 
	Even with our proposed voxel query,
	the local aggregation operation (\emph{i.e.}, PointNet module) in Voxel RoI pooling still involves large computation complexity. 
	As shown in Figure \ref{fig:mlp} (a), there are totally $M$ grid points ($M=r\times G^3$, where $r$ is the number of RoI, and $G$ is the grid size), and $K$ voxels are grouped for each grid point. The dimension of grouped feature vectors is $C+3$, including the C-dim voxel features and 3-dim relative coordinates. 
	The grouped voxels occupy a lot of memories and lead to large computation FLOPs ($O(M\times K \times(C+3)\times C')$) when applying the FC layer. 
	
	As motivated by \shortcite{liu2020closer, hu2020randla}, we additionally introduce an accelerated PointNet Module to further reduce the computation complexity of Voxel Query. Typically, as shown in Figure \ref{fig:mlp} (b), the voxel features and relative coordinates are decomposed into two streams.  Given the FC layer with weight $W\in\mathds{R}^{C',C+3}$, we divide it into $W_\text{F}\in\mathds{R}^{C',C}$ and $W_\text{C}\in\mathds{R}^{C',3}$. Since the voxel features are independent of the grid points, we apply a FC layer with $W_\text{F}$ on the voxel features before performing voxel query. Then, after voxel query, we only multiply the grouped relative coordinates by $W_\text{C}$ to obtain the relative position features and add them to the grouped voxel features. The FLOPs of our accelerated PointNet module are $O(N\times C \times C'+M\times K\times 3\times C')$. Since the number of grouped voxels $(M\times K)$ is an order of magnitude higher than $N$, the accelerated PointNet module is more efficient than the original one.

	\subsection{Backbone and Region Proposal Networks} 
	We follow the similar design of \shortcite{yan2018second,He_2020_CVPR,Shi_2020_CVPR} to build our backbone networks. The 3D backbone network gradually converts the voxelized inputs into feature volumes.
	Then, the output tensors are stacked along the Z axis to produce BEV feature maps. The 2D backbone network consists of two components: a top-down feature extraction sub-network with two blocks of standard $3\times3$ convolution layers and a mutli-scale feature fusion sub-network that upsamples and concatenates the top-down features. Finally, the output of the 2D backbone network is convolved with two sibling $1\times1$ convolutional layers to generate 3D region proposals. We will detail the architecture of our backbone networks in Section 4.2.
	
	\subsection{Detect Head} 
	The detect head takes RoI features as input for box refinement. Specifically, 
	a shared 2-layer MLP first transform RoI features into feature vectors. Then, the flattened features are injected into two sibling branches: one for bounding box regression and the other for confidence prediction. As motivated by \shortcite{jiang2018acquisition,li2019gs3d,Shi_2020_CVPR,shi2020points}, the box regression branch predicts the residue from 3D region proposals to the ground truth boxes, and the confidence branch predicts the IoU-related confidence score.
	
	\subsection{Training Objectives} 
	
	\subsubsection{Losses of RPN.} \hskip -5pt We follow \shortcite{yan2018second,lang2019pointpillars} to devise the losses of the RPN as a combination of classification loss and box regression loss, as:
	\begin{equation}\label{equ.loss_rpn}\small
		\mathcal{L}_\text{RPN} = \frac{1}{N_{\text{fg}}}[\sum_{i} \mathcal{L}_{\text{cls}}(p_i^a, c_i^{\ast})
		+\mathds{1}(c_i^{\ast}\ge{1})\sum_{i} \mathcal{L}_{\text{reg}}(\delta_i^a, t_i^{\ast})],
	\end{equation}
	where $N_\text{fg}$ represents the number of foreground anchors, $p_i^a$ and $\delta_i^a$ are the outputs of classification and box regression branches, $c_i^{\ast}$ and $t_i^{\ast}$ are the classification label and regression targets respectively. $\mathds{1}(c_i^{\ast}\ge{1})$ indicates regression loss in only calculated with foreground anchors. Here we utilize Focal Loss \shortcite{lin2017focal} for classification and Huber Loss for box regression. 
	
	\subsubsection{Losses of detect head.} \hskip -5pt The target assigned to the confidence branch is an IoU related value, as:
	\begin{equation}
		l_i^\ast(\text{IoU}_i)=\left\{
		\begin{array}{lll}
			0 & & {\text{IoU}_i < \theta_L},\\
			\frac{\text{IoU}_i - \theta_L}{\theta_H-\theta_L} & & {\theta_L \leq \text{IoU}_i < \theta_H},\\
			1 & & {\text{IoU}_i > \theta_H},
		\end{array} \right.
	\end{equation}
	where $\text{IoU}_i$ is the IoU between the $i$-th proposal and the corresponding ground truth box, $\theta_H$ and $\theta_L$ are foreground and background IoU thresholds. Binary Cross Entropy Loss is exploited here for confidence prediction. The box regression branch also uses Huber Loss as in the RPN. The losses of our detect head are computed as:
	\begin{equation}\label{equ.loss_head}\small
		\begin{split}
			\mathcal{L}_\text{head} = &\frac{1}{N_s}[\sum_{i} \mathcal{L}_{\text{cls}}(p_i,l_i^\ast(\text{IoU}_i))\\
			&+\mathds{1}(\text{IoU}_i\ge{\theta_{reg}})\sum_{i} \mathcal{L}_{\text{reg}}(\delta_i, t_i^{\ast})],
		\end{split}
	\end{equation}
	where $N_s$ is the number of sampled region proposals at the training stage, and $\mathds{1}(\text{IoU}_i\ge{\theta_{reg}})$ indicates that only region proposals with $\text{IoU} > \theta_\text{reg}$ contribute to the regression loss.

	\section{Experiments}
	
	\subsection{Datasets}
	
	\subsubsection{KITTI.} \hskip -5pt
	The \textit{KITTI Dataset} \shortcite{geigerwe} contains 7481 training samples and 7518 testing samples in autonomous driving scenes. As a common practice, the training data are divided into a \textit{train} set with $3712$ samples and a \textit{val} set with $3769$ samples.  The performance on both the \textit{val} set and online test leaderboard are reported for comparison. When performing experimental studies on the \textit{val} set, we use the \textit{train} set for training. For test server submission, we randomly select $80\%$ samples from the training point clouds for training, and use the remaining $20\%$ samples for validation as in \shortcite{Shi_2020_CVPR}.

	\subsubsection{Waymo Open Dataset.} \hskip -5pt
	The \textit{Waymo Open Dataset} \shortcite{sun2020scalability} is the largest public dataset for autonomous driving to date. There are totally $1,000$ sequences in this dataset, including $798$ sequences ($\sim$$158\text{k}$ point clouds samples) in training set and 202 sequences ($\sim$$40\text{k}$ point clouds samples) in validation set. Different from KITTI that only provides annotations in camera FOV, the 
	\textit{Waymo Open Dataset} provides annotations for objects in the full $360^\circ$.
	
	\subsection{Implementation Details}
	
	\subsubsection{Voxelization.} \hskip -5pt 
	The raw point clouds are divided into regular voxels before taken as the input of our Voxel R-CNN. Since the KITTI Dataset only provides the annotations of object in FOV, we clip the range of point clouds into $[0, 70.4]m$ for the $X$ axis, $[-40,40]m$ for the $Y$ axis and $[-3,1]m$ for Z axis. The input voxel size is set as $(0.05m, 0.05m, 0.1m)$. For the Waymo Open Dataset, the range of point clouds is clipped into $[-75.2, 75.2]m$ for X and Y axes and $[-2, 4]m$ for the Z axis. The input voxel size is set as $(0.1m, 0.1m, 0.15m)$.

	\subsubsection{Network Architecture.} \hskip -5pt
	The architecture of 3D backbone and 2D backbone follows the design in \cite{yan2018second,Shi_2020_CVPR}. There are four stages in the 3D backbone with filter numbers of $16, 32, 48, 64$ respectively. There are two blocks in the 2D backbone network, the first block keeps the same resolution along $X$ and $Y$ axes as the output of 3D backbone network, while the second block is half the resolution of the first one. The numbers of convolutional layers in these two blocks, \emph{i.e.}, $N1$ and $N2$, are both set as 5. And the feature dimensions of these blocks are $(64, 128)$ for KITTI Dataset and $(128, 256)$ for Waymo Open Dataset.
	Each region proposal is divided into a $6\times 6\times 6$ grid for Voxel RoI pooling. The Manhattan distance thresholds are set as $2$ and $4$ for multi-scale voxel query. The output channels of the local feature aggregation module are $32$ for KITTI Dataset and $64$ for Waymo Open Dataset.

		\begin{table}
		\small
		\renewcommand\arraystretch{1.05}
		\begin{center}
			\scalebox{0.8}[0.8]{
				\setlength\tabcolsep{8pt}
				\begin{tabular}{c|c|ccc}
					\hline
					\multirow{2}{*}{Method} &
					\multicolumn{1}{c|}{FPS} &  
					\multicolumn{3}{c}{$\text{AP}_\text{3D}$ (\%) ~~} \\
					&(Hz)&Easy & Mod. & Hard \\
					\hline
					\multicolumn{5}{c}{RGB+LiDAR} \\
					\hline
					MV3D \shortcite{chen2017multi} & - & 74.97 & 63.63 & 54.00 \\
					AVOD-FPN \shortcite{ku2018joint} &  10.0 &  83.07 & 71.76 & 65.73 \\   
					F-PointNet \shortcite{qi2018frustum} &  5.9 & 82.19 & 69.79 & 60.59 \\
					PointSIFT+SENet \shortcite{zhao20193d} &  - & 85.99 & 72.72 & 64.58 \\
					UberATG-MMF \shortcite{liang2019multi} & - & 88.40 & 77.43 & 70.22 \\
					\hline
					\multicolumn{5}{c}{LiDAR-only} \\
					\hline
					\textbf{Point-based:} &  &  &  &  \\
					PointRCNN \shortcite{shi2019pointrcnn} &10.0 & 86.96 & 75.64 & 70.70  \\
					STD \shortcite{Yang_2019_ICCV} &12.5 &  87.95 & 79.71 & 75.09\\
					Patches \shortcite{lehner2019patch} &- &  88.67 & 77.20 & 71.82 \\
					3DSSD \shortcite{Yang_2020_CVPR} &  26.3 &88.36 & 79.57 & 74.55\\
					PV-RCNN \shortcite{Shi_2020_CVPR} &8.9 &  90.25 & 81.43 & 76.82 \\
					\textbf{Voxel-based:} &  &  &  &  \\
					VoxelNet \shortcite{zhou2018voxelnet} & - & 77.47 & 65.11 & 57.73 \\
					SECOND \shortcite{yan2018second} &   30.4 & 83.34 & 72.55 & 65.82 \\
					PointPillars \shortcite{lang2019pointpillars} &  42.4 & 82.58 & 74.31 & 68.99 \\
					Part-$A^2$ \shortcite{shi2020points} & - & 87.81 & 78.49 & 73.51 \\
					TANet \shortcite{liu2020tanet} &  28.7 & 85.94 & 75.76 & 68.32 \\
					HVNet \shortcite{Ye_2020_CVPR} & 31.0 &  87.21 & 77.58 & 71.79 \\
					SA-SSD \shortcite{He_2020_CVPR} & 25.0 &  88.75 & 79.79 & 74.16\\
					\hline\hline
					Voxel R-CNN (ours) & 25.2 & \textbf{90.90} & \textbf{81.62} & \textbf{77.06} \\
					\hline
				\end{tabular}
			}
		\end{center}
		\caption{Performance comparison on the KITTI \textit{test} set with AP calculated by recall 40 positions for car class}
		\label{tab:test}
	\end{table}

	\subsubsection{Training.} \hskip -5pt
	The whole architecture of our Voxel R-CNN is end-to-end optimized with the Adam optimizer.  For KITTI Dataset, the network is trained for 80 epochs with the batch size 16. For Waymo Open Dataset, the network is trained for 30 epochs with the batch size 32. The learning rate is initialized as $0.01$ for both datasets and updated by cosine annealing strategy.
	In the detect head, the foreground IoU threshold $\theta_\text{H}$ is set as 0.75,  background IoU threshold $\theta_\text{L}$ is set as 0.25, and the box regression IoU threshold $\theta_\text{reg}$ is set as 0.55. We randomly sample $128$ RoIs as the training samples of detect head. Within the sampled RoIs, half of them are positive samples that have IoU $>$ $\theta_\text{reg}$ with the corresponding ground truth boxes. We conduct data augmentation at the training stage following 
	strategies in \shortcite{lang2019pointpillars,Shi_2020_CVPR,Yang_2020_CVPR,Ye_2020_CVPR}. Please refer to OpenPCDet \shortcite{openpcdet2020} for more detailed configurations since we conduct all experiments with this toolbox.

	\subsubsection{Inference.} \hskip -5pt
	At the inference stage, we first perform non-maximum suppression (NMS) in the RPN with IoU threshold $0.7$ and keep the top-100 region proposals as the input of detect head. Then, after refinement, NMS is applied again with IoU threshold $0.1$ to remove the redundant predictions.

	\subsection{Results on KITTI Dataset}
	We evaluate our Voxel R-CNN on KITTI Dataset following the common protocol to report the average precision (AP) of class Car with the 0.7 (IoU) threshold. We report our performance on both $val$ set and $test$ set for comparison and analysis. The performance on \textit{val} set is calculated with the AP setting of recall 11 positions. And the results evaluated by the test server utilize AP setting of recall 40 positions$\footnote{The setting of AP calculation is modified from recall 11 positions to recall 40 positions on 08.10.2019. We exploits the recall 11 setting on $val$ set for fair comparison with previous methods.}$ .

	\begin{table}[!t]
	\small 
	\begin{center}
		\scalebox{0.99}[0.9]{
			\setlength\tabcolsep{4.2pt}
			\begin{tabular}{c|ccc|ccc}
				\hline 
				\multirow{2}{*}{\tabincell{c}{IoU \\ Thresh.}} & 
				\multicolumn{3}{c|}{$\text{AP}_\text{3D}$ (\%)} & 
				\multicolumn{3}{c}{$\text{AP}_\text{BEV}$ (\%)}\\
				& Easy & Moderate  & Hard & Easy & Moderate  & Hard \\
				\hline
				0.7 & 92.38 & 85.29 & 82.86 & 95.52 & 91.25 & 88.99 \\ 
				\hline
			\end{tabular}
		}
	\end{center}
	\caption{Performance of Voxel R-CNN on the KITTI \emph{val}  set with AP calculated by 40 recall positions for car class}
	\label{tab:val_r40}
\end{table}

	\begin{table}[!t]
		\small
		\renewcommand\arraystretch{1}
		\begin{center}
			\scalebox{0.75}[0.75]{
				\setlength\tabcolsep{11pt}
				\begin{tabular}{c|ccc}
					\hline
					\multirow{2}{*}{Method} &
					\multicolumn{3}{c}{$\text{AP}_\text{3D}$ (\%) ~~} \\
					&Easy & Moderate & Hard \\
					\hline
					\textbf{Point-based:} & &  &   \\
					PointRCNN \shortcite{shi2019pointrcnn} & 88.88 &78.63& 77.38  \\
					STD \shortcite{Yang_2019_ICCV}  &  89.70 & 79.80 & \textbf{79.30}\\
					3DSSD \shortcite{Yang_2020_CVPR}  &89.71 & 79.45 & 78.67\\
					PV-RCNN \shortcite{Shi_2020_CVPR} &  89.35 & 83.69 & 78.70 \\
					\textbf{Voxel-based:} & &  &   \\
					VoxelNet \shortcite{zhou2018voxelnet} & 81.97 & 65.46 & 62.85 \\
					SECOND \shortcite{yan2018second}  & 88.61 & 78.62 & 77.22 \\
					PointPillars \shortcite{lang2019pointpillars} &86.62 &76.06 &68.91  \\
					Part-$A^2$ \shortcite{shi2020points} & 89.47 & 79.47 & 78.54 \\
					TANet \shortcite{liu2020tanet}  & 87.52 & 76.64 & 73.86 \\
					
					SA-SSD \shortcite{He_2020_CVPR}  &  \textbf{90.15} & 79.91 & 78.78\\
					\hline\hline
					Voxel R-CNN (ours)  & 89.41 & \textbf{84.52} & 78.93 \\
					\hline
				\end{tabular}
			}
		\end{center}
		\caption{Performance comparison on the KITTI \textit{val} set with AP calculated by 11 recall positions for car class
		}
		\label{tab:val_r11}
	\end{table}

	\subsubsection{Comparison with State-of-the-art Methods.} \hskip -5pt
	We compare our Voxel R-CNN with several state-of-the-art methods on the KITTI $test$ set by submitting our results to the online test server. As shown in the Table \ref{tab:test}, our Voxel R-CNN makes the best balance between the accuracy and efficiency among all the methods. By taking full advantage of voxel-based representation, Voxel R-CNN achieves $81.62\%$ average precision (AP) on the moderate level of class Car, with the real time processing frame rate ($25.2$ FPS). Particularly, Voxel R-CNN achieves comparable accuracy with the strongest competitor, \emph{i.e.}, PV-RCNN \shortcite{Shi_2020_CVPR}, with only about $1/3$ running time. This verifies the  spatial  context  in  the  voxel  representation  is  almost sufficient for 3D object detection, and the voxel representation is more efficient for feature extraction. Besides, Voxel R-CNN outperforms all of the existing voxel-based models by a large margin, \emph{i.e.}, $2.15\%, 1.83\%, 2.90\%$ absolute improvements for easy, moderate and hard level over the SA-SSD\shortcite{He_2020_CVPR}. 

	The AP for 3D object detection and BEV object detection of our Voxel R-CNN on KITTI Dataset is presented in Table \ref{tab:val_r40}. The results in this table are calculated by recall 40 positions with the (IoU) threshold $0.7$. In addition, we report the performance on the KITTI $val$ set with AP calculated by recall 11 positions for further comparison. As shown in Table \ref{tab:val_r11}, our Voxel R-CNN achieves the best performance on moderate and hard level on the $val$ set.
	
	Overall, the results on both $val$ set and $test$ set consistently demonstrate that our proposed Voxel R-CNN achieves the state-of-the-art average precision on 3D object detection and keeps the high efficiency of voxel-based models.

\begin{table}
	\small
	\begin{center}
		\scalebox{0.72}[0.72]{
			\setlength\tabcolsep{5pt}
			\begin{tabular}{c|cccc}
				\hline
				Method
				& Overall & 0-30m & 30-50m & 50m-Inf \\
				\hline
				\textbf{\textit{LEVEL\_1 3D mAP (IoU=0.7):}} & &  & \\
				PointPillar \shortcite{lang2019pointpillars} & 56.62 & 81.01 & 51.75 & 27.94  \\
				MVF \shortcite{zhou2020end} & 62.93 & 86.30 & 60.02 & 36.02 \\
				Pillar-OD \shortcite{wang2020pillar} & 69.80 & 88.53 & 66.50 & 42.93 \\
				AFDet+PointPillars-0.10 \shortcite{ge2020afdet} & 63.69 & 87.38 & 62.19 & 29.27 \\
				PV-RCNN \shortcite{Shi_2020_CVPR}& {70.30} & {91.92} & {69.21} & {42.17} \\
				\textbf{Voxel R-CNN (ours)} &\textbf{75.59}& \textbf{92.49}&\textbf{74.09}&\textbf{53.15} \\
				\hline 
				\textbf{\textit{LEVEL\_1 BEV mAP (IoU=0.7):}} & &  &   \\
				PointPillar \shortcite{lang2019pointpillars} & 75.57 & 92.1 & 74.06 & 55.47 \\ 
				MVF \shortcite{zhou2020end} & 80.40 & 93.59 & 79.21 & 63.09 \\
				Pillar-OD \shortcite{wang2020pillar} & 87.11 & 95.78 & 84.87 & 72.12 \\				
				PV-RCNN \shortcite{Shi_2020_CVPR}& {82.96} & {97.35} & {82.99} & {64.97} \\ 
				\textbf{Voxel R-CNN (ours)} & \textbf{88.19} & \textbf{97.62} & \textbf{87.34} & \textbf{77.70}  \\
				\hline
				\textbf{\textit{LEVEL\_2 3D mAP (IoU=0.7):}} & &  & \\
				PV-RCNN \shortcite{Shi_2020_CVPR}& 65.36 & 91.58 & 65.13 & 36.46 \\
				\textbf{Voxel R-CNN (ours)} &\textbf{66.59} & \textbf{91.74} & \textbf{67.89} & \textbf{40.80} \\
				
				\hline 
				\textbf{\textit{LEVEL\_2 BEV mAP (IoU=0.7):}} & &  & \\
				PV-RCNN \shortcite{Shi_2020_CVPR}& 77.45 & 94.64 & 80.39 & 55.39 \\
				\textbf{Voxel R-CNN (ours)} &  \textbf{81.07} & \textbf{96.99} & \textbf{81.37} & \textbf{63.26} \\
				\hline 
			\end{tabular}
		}
	\end{center}
	\caption{Performance comparison on the Waymo Open Dataset with 202 validation sequences ($\sim$40k samples) for the vehicle detection
	}
	\label{tab:waymo}
\end{table}
	
	\subsection{Results on Waymo Open Dataset}
	We also conduct experiments on the larger Waymo Open Dataset to further validate the effectiveness of our proposed Voxel R-CNN. The objects on the Waymo Open Dataset are split into two levels based on the number of points of a single object, where the LEVEL\_1 objects have more than 5 points while the LEVEL\_2 objects have 1$\sim$5 points.

	\subsubsection{Comparison with State-of-the-art Methods.} \hskip -5pt
	We evaluate our Voxel R-CNN on both LEVEL\_1 and LEVEL\_2 objects and compare with several top-performing methods on the Waymo Open Dataset. Table~\ref{tab:waymo} shows that our method surpasses all previous methods with remarkable margins on all ranges of both LEVEL\_1 and LEVEL\_2. Specifically, with the commonly used LEVEL\_1 3D mAP evaluation metric, our Voxel R-CNN achieves new state-of-the-art performance with 75.59\% mAP, and outperforms previous state-of-the-art method PV-RCNN by 5.29\% mAP. Especially, our method outperforms PV-RCNN with +10.98\% mAP in the range of 50m-Inf, the significant gains on the far away area demonstrate the effectiveness of our proposed Voxel R-CNN for detecting the objects with very sparse points.

	\subsection{Ablation Study}

	Table \ref{tab:ablation} details how each proposed module influences the accuracy and efficiency of our Voxel R-CNN. The results are evaluated with AP of moderate level for car class.
	
	\textit{Method (a)} is the one stage baseline that performs detection on BEV features. It runs at 40.8 FPS, but with unsatisfactory AP, which indicates \textit{BEV representation alone is insufficient to precisely detect objects in a 3D space}.
	
	\textit{Method (b)} extends \textit{(a)} with a detect head for box refinement, which leads to a boost of $4.22\%$ moderate AP. This verifies \textit{the spatial context from 3D voxels provides sufficient cues for precise object detection}. However, \textit{(b)} applies time-consuming ball query and original PointNet modules to extract RoI features, leading to a decrease of 23.4 FPS.
	
	\textit{Method (c)} replaces ball query with our voxel query, which makes 1.7 FPS improvement by \textit{taking advantage of the neighbor-aware property of voxel-based representations}.
	
	\textit{Method (d)} uses our accelerated PointNet module to aggregate voxel features, boosting the FPS from $17.4$ to $21.4$.
	
	\textit{Method (e)} is the proposed Voxel R-CNN. By extending the one stage baseline with a detect head, taking voxel query for voxel grouping, and utilizing the accelerated PointNet module for local feature aggregation, Voxel R-CNN \textit{achieves the state-of-the-art accuracy} for 3D object detection and \textit{maintain the high efficiency} of voxel-based models.
	
	\begin{table}
	\renewcommand\arraystretch{1}
	\small
	\begin{center}
		\scalebox{0.9}[0.9]{
			\setlength\tabcolsep{6pt}
			\begin{tabular}{c|ccc|c|c}
				\hline
				Methods &D.H. & V.Q.  & A.P. & Moderate $\text{AP}_\text{3D}$ (\%)& FPS (Hz) \\
				\hline
				(a) & &  &    & 77.99  & 40.8 \\
				(b) &  $\bm{\surd}$   &  &   & 84.21  & 17.4 \\
				(c) &  $\bm{\surd}$  & $\bm{\surd}$  &  & 84.33  &  19.1\\
				(d) & $\bm{\surd}$   &  & $\bm{\surd}$  & 84.27  & 21.4  \\
				(e) & $\bm{\surd}$   & $\bm{\surd}$& $\bm{\surd}$  & 84.52 & 25.2 \\
				\hline
			\end{tabular}
		}
	\end{center}
	\caption{Performance of proposed method with different configurations on KITTI \textit{val} set. ``D.H.'',  ``V.Q.'' and ``A.P.'' stand for detect head, voxel query and accelerated PointNet module respectively. The results are evaluated with the average precision calculated by 11 recall positions for car class.
	}
	\label{tab:ablation}
\end{table}

	\section{Related Work}
	We broadly categorize the methods for 3D object detection on point clouds into point-based \shortcite{shi2019pointrcnn,Yang_2019_ICCV,Chen2019fastpointrcnn,Yang_2020_CVPR,Shi_2020_CVPR} methods and voxel-based \shortcite{zhou2018voxelnet,yan2018second,yang2018pixor,simon2018complex,lang2019pointpillars,liu2020tanet} methods. Here we give a brief review on these two directions:
	
	\textbf{Point-based Methods.}  Point-based methods take the raw point clouds as inputs, and abstract a set of point representations with iterative sampling and grouping \shortcite{qi2017pointnet,qi2017pointnet++}.
	PointRCNN \shortcite{shi2019pointrcnn}  introduces a 3D region proposal network based on PointNet++ \shortcite{qi2017pointnet++} and utilizes point cloud RoI pooling to extract 3D region features of each proposal. STD \shortcite{Yang_2019_ICCV} exploits the PointsPool operation to extract points features of each proposal and transfer the sparse points into dense voxel representation. Then, 3D CNNs are applied on the voxelized region features for further refinement. 3DSSD \shortcite{Yang_2020_CVPR} introduces F-FPS as a supplement of D-FPS and builds a one stage anchor-free 3D object detector based on feasible representative points. Different from grouping neighboring points for set abstraction, PV-RCNN \shortcite{Shi_2020_CVPR} devises voxel set abstraction to integrates multi-scale voxel features into sampled keypoints.  By leveraging both the accurate position information from raw points and spatial context from voxel representations, PV-RCNN achieves  remarkable improvements on 3D object detection.

	\textbf{Voxel-based Methods.} Voxel-based methods typically discrete point clouds into equally spaced grids, and then capitalize on 2D/3D CNN to perform object detection. The early work VoxelNet \shortcite{zhou2018voxelnet} first divides points into 3D voxels and uses a tiny PointNet to transform points of each voxel into a compact feature representation. Then, 3D CNNs are leveraged to aggregate spatial context and generate 3D detections. However, due to the sparsity of non-empty voxels in the large space, it is inefficient to exploit conventional convolution networks for feature extraction.

	To reduce the computation cost,  SECOND \shortcite{yan2018second} introduces sparse 3D convolution for efficient voxel processing. SA-SSD \shortcite{He_2020_CVPR} proposes an auxiliary network and losses on the basis of SECOND \shortcite{yan2018second} to preserve structure information. Different from the methods operating on voxels in 3D space, 
	PointPillars \shortcite{lang2019pointpillars} groups points as ``pillars'' and capitalizes on a simplified PointNet \shortcite{qi2017pointnet,qi2017pointnet++} for pillar feature extraction before forming pseudo BEV images.

	\section{Conclusion}
	In this paper, we present Voxel R-CNN, a novel 3D object detector with voxel-based representations. Taking the voxels as input, Voxel R-CNN first generates dense region proposals from bird-eye-view feature representations, and subsequently, utilizes voxel RoI pooling to extract region features from 3D voxel features for further refinement. By taking full advantage of voxel representations, our Voxel R-CNN strikes a careful balance between accuracy and efficiency. The encouraging results on both KITTI Dataset and Waymo Open Dataset demonstrate our Voxel-RCNN can serve as a simple but effective baseline to facilitate the investigation of 3D object detection and other downstream tasks.
	
	\section{Acknowledgements}
	This work was supported by the National Major Program for Technological Innovation 2030—New Generation Artificial Intelligence (2018AAA0100500). The work of Wengang Zhou was supported in part by NSFC under Contract 61822208, 61632019 and 62021001, and in part by the Youth Innovation Promotion Association CAS under Grant 2018497. The work of Houqiang Li was supported by NSFC under Contract 61836011.

	\newpage
	\bibliographystyle{aaai}
	\bibliography{refs.bib}

\end{document}